\documentclass[final]{cvpr}

\usepackage{times}
\usepackage{epsfig}
\usepackage{graphicx}
\usepackage{amsmath}
\usepackage{amssymb}
\usepackage{bm}
\usepackage[shortlabels]{enumitem}
\usepackage{breqn}
\usepackage{verbatim}
\usepackage{tabularx}

\usepackage{color}

\newtheorem{theorem}{Theorem}

\newtheorem{definition}[theorem]{Definition}

\newtheorem{remark}{Remark}

\numberwithin{theorem}{section}
\usepackage{bbding}

\usepackage[pagebackref=true,breaklinks=true,colorlinks,bookmarks=false]{hyperref}

\def\cvprPaperID{5163} 
\def\confYear{CVPR 2021}

\begin{document}

\title{Causal Hidden Markov Model for Time Series Disease Forecasting}


\author{Jing~Li$^{1,2}$,\quad Botong~Wu$^{1,5}$, \quad Xinwei~Sun$^{4}$\Envelope{},  
\quad  Yizhou~Wang $^{1,3}$
\\
\textsuperscript{1} Dept. of  Computer Science, Peking University \quad
\textsuperscript{2} Adv. Inst. of Info. Tech, Peking University
\\
\textsuperscript{3} Center on Frontiers of Computing Studies, Peking University \\
\textsuperscript{4} Microsoft Research, Asia \quad
\textsuperscript{5} Deepwise AI Lab \\
{\tt\small \tt\small {\{lijingg, botongwu, yizhou.wang\}}@pku.edu.cn,  xinsun@microsoft.com}


}

\maketitle

\begin{abstract}
We propose a causal hidden Markov model to achieve robust prediction of irreversible disease at an early stage, which is safety-critical and vital for medical treatment in early stages. Specifically, we introduce the hidden variables which propagate to generate medical data at each time step. To avoid learning spurious correlation (\textit{e.g.}, confounding bias), we explicitly separate these hidden variables into three parts: a) the disease (clinical)-related part; b) the disease (non-clinical)-related part; c) others, with only a),b) causally related to the disease however c) may contain spurious correlations (with the disease) inherited from the data provided. With personal attributes and disease label respectively provided as side information and supervision, we prove that these disease-related hidden variables can be disentangled from others, implying the avoidance of spurious correlation for generalization to medical data from other (out-of-) distributions. Guaranteed by this result, we propose a sequential variational auto-encoder with a reformulated objective function. We apply our model to the early prediction of peripapillary atrophy and achieve promising results on out-of-distribution test data. Further, the ablation study empirically shows the effectiveness of each component in our method. And the visualization shows the accurate identification of lesion regions from others. 
\footnote{This project is released on: \sf{https://sites.google.com/view/causal-hmm}}

\end{abstract}

\section{Introduction}




Future disease forecasting is especially important for those irreversible diseases, such as Alzheimer's Disease \cite{sun2017gsplit} and eye diseases \cite{carr2017science}. Forecasting at an early stage provides the doctors with a window to implement medical treatment/intervention such as drug or physical exercise, in order to slow down or alleviate the disease progression. However, such early forecasting can face the following challenges: i) the incomplete information for forecasting, \emph{i.e.}, the lack of medical observation of the future stage; ii) the medical data (such as images and clinical measurements) can suffer from distributional change across populations or hospitals, the forecasting on which is known as out-of-distribution (OOD) generalization that can fail many existing supervised learning models (such as adversarial attack \cite{goodfellow2014explaining}).



Existing works for future disease forecasting can be roughly categorized into two classes: 1) forecasting with additional supervisions; 2) forecasting based on generation of future image. For the first class(\textit{e.g.}, \cite{7452361,ning2020ldgan,Jiang2018ophthalmic,TABRIZI201131}), this additional supervision can refer to disease labels at each time step or the target image at future stage. Therefore, they can not be adopted to scenarios when these supervisions are lacked (\textit{e.g.}, the disease label at the current stage is unprovided due to labeling cost). For the second class(\textit{e.g.}, \cite{louis2019riemannian}, they aim to exploit latent space to generate the sequential medical images, which can be followed by a disease classifier for disease prediction. Although these methods can achieve accurate generation \cite{louis2019riemannian}, they may suffer from learning spurious correlated features due to biases inherited from the data provided. These biases can refer to correlated but disease-unrelated information such as background or clinical attributes, which are data-dependent and hence may not be robust to other data distributions (\textit{i.e.}, out-of-distribution).

\begin{figure*}[tb]
\centering
\includegraphics[width=0.9\linewidth]{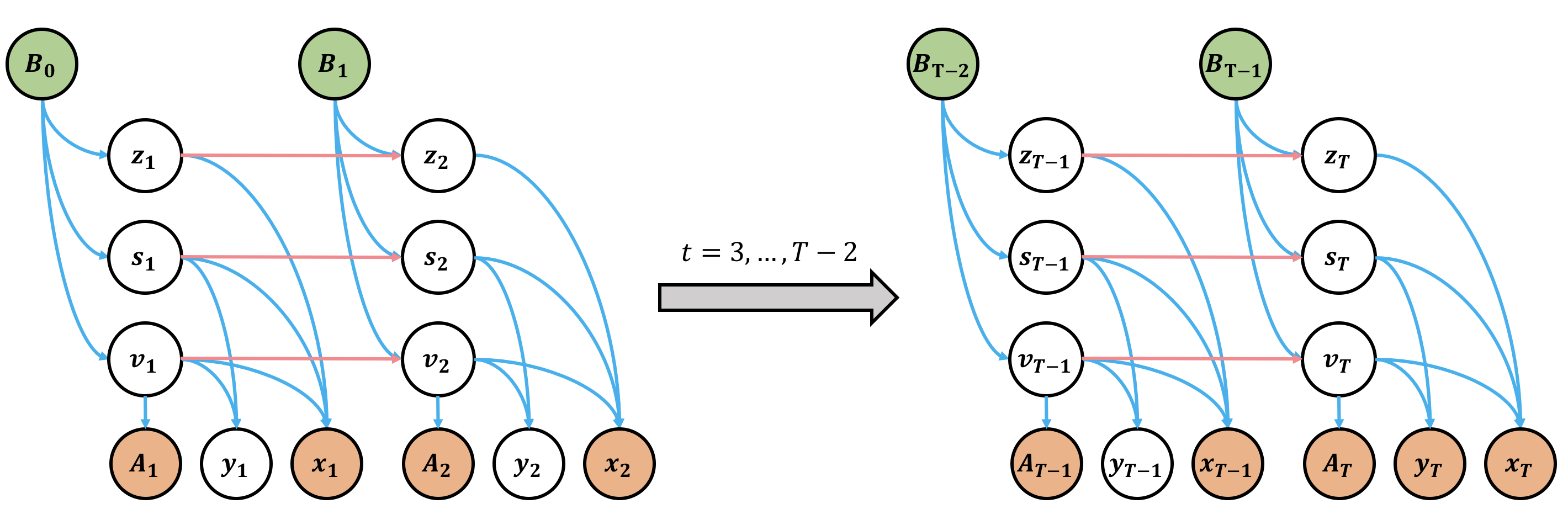}
\caption{The directed acylic graph (DAG) for our Causal-HMM. The hidden variables independently propagate as $t$ grows. At each step, the personal attributes $\bm{B}_{t-1}$ generate the image $\bm{x}_t$'s latent components $\bm{s}_t,\bm{v}_t,\bm{z}_t$, among which $\bm{v}_t$ points to the $\bm{A}_t$ and $\bm{v}_t,\bm{s}_t$ point to $y_t$. At final stage $T$, the disease label $y_T$ is causally related to $\bm{s}_T,\bm{v}_T$. Note that colorless variable means unobserved variable, colored variable means observed one.} 
\label{fig:framework}
\end{figure*}

To avoid spurious correlation to enable robust forecasting, we propose \textbf{Causal} \textbf{H}idden \textbf{M}arkov \textbf{M}odel (Causal-HMM) in which we explicitly separate the disease-causative features from others, and model them using hidden variables that propagate to generate the medical observation, as encapsulated in the causal graph in Fig.~\ref{fig:framework}. Specifically as shown, among all hidden variables, \textit{i.e.}, $\bm{s},\bm{v},\bm{z}$ that generate the medical image $\bm{x}$, only $\bm{s},\bm{v}$ are causally related to the disease label $y$. Taking peripapillary atrophy disease \cite{10.1167/iovs.12-9682} as an example, the $\bm{v}$ is related to the clinical measurements that are relevant to the disease (denoted as $\bm{A}$, \textit{e.g.}, Axial length, Corneal thickness, Corneal curvature) and the $\bm{s}$ is related to other disease-related aspects that beyond clinical measurements but can be reflected in the retinal image (\textit{e.g.}, Maculopathy, Fundus morphology, choroidal vessels). The propagation of these hidden variables are confounded by personal attributes (denoted as $\bm{B}$ such as \textit{age, gender, etc.}), making the $\bm{z}$ spuriously correlated with the disease and can be learned according to intrinsic bias from observed data. We theoretically show that the hidden variables (\textit{i.e.}, $\bm{s}$, $\bm{v}$) can be possibly identified from observational distribution, benefited from the explicit separation of $\bm{s}$, $\bm{v}$ and $\bm{z}$. To the best of our knowledge, we are \textit{the first} to provide the identifiability result for sequential data in the supervised scenario. For practical inference, we reformulate a new sequential VAE framework that conforms to the causal model above. The disentangled disease-related hidden variables are used for future disease prediction.


We apply our method on an in-house dataset of peripapillary atrophy (PPA) which is related to many irreversible eye diseases. The dataset is divided into training set, validation set and test set with the first two share the same distribution. The empirical results show that compared with other methods, our Causal-HMM can achieve much better prediction accuracy on the out-of-distribution test set, implying the ability to handle spurious correlation of our method. The ablative studies show the effectiveness of each component in our method. The visualization further shows that the identified disease-causative hidden variables by our model are concentrated on the disease-related regions. Our contribution can be summarized as follows: 

\begin{itemize}
    \item \textbf{Methodologically}, we propose a novel Causal Hidden Markov Model of sequential data for future disease forecasting;
    \item \textbf{Algorithmically}, we reformulate a new sequential VAE framework, which is aligned with the causal model above;
    \item \textbf{Theoretically}, we provide an identifiability result, which implicitly ensures disentanglement of the disease-causative latent features from others;
    \item \textbf{Experimentally}, we achieve SOTA prediction result for the peripapillary atrophy forecasting problem, on the out-of-distribution test dataset. 
\end{itemize}

\section{Related Work}
Conventional methods for disease progression require additional data, which can refer to the supervision signals (\emph{e.g.,} disease labels) at each time step, or target information at the future stage (\emph{e.g.,} image or clinical measurements at future stage), \emph{e.g.,} \cite{7452361,ning2020ldgan,Jiang2018ophthalmic,TABRIZI201131}. Different from them, in our disease forecasting scenario, the target information cannot be observed. Besides, the disease label at each time step is also not provided, which is common in many practical medical scenarios due to large labeling costs. 

Other works with more similar settings to ours are modeling sequential or time series data, \emph{e.g.,} \cite{louis2019riemannian,Gao2018BrainDD,8363833,li2019enhancing}. Most of these work implemented Multilayer Perception (MLP) or Convolutional neural networks (CNN) for feature extraction; and use a recurrent neural network (RNN) to generate the trajectory of extracted features, \emph{e.g.,}\cite{Gao2018BrainDD} and \cite{8363833}.
Instead the \cite{li2019enhancing} proposes to use transformer to capture the long term dependencies.
Particularly, the \cite{louis2019riemannian} proposed a deep generative model that learned from the low dimensional latent space that assumed to lie on an priori known Riemannian manifold. Specifically, it implemented an RNN to encode the sequential image into hidden embedding and then decoded them to generate the sequential data. However, these models do not separate the features that are causally related to the disease label from others, making them suffer from learning spurious correlation. This spurious correlation inherited from data may not hold on OOD samples, which can cause a high risk for the safety-critical medical data. \textbf{In contrast}, our method explicitly disentangle these disease-causative features (\textit{i.e.}, the hidden variables that are related to the disease) from others, in order to avoid spurious correlation and hence enable the OOD generalization. 


There are also works for learning disentanglement of latent space, \emph{i.e.,}\cite{ilse2020diva,mahajan2020diverse}. Specifically, DIVA~\cite{ilse2020diva} proposes a generative model by learning three subspaces that account for domain, class and residual variations respectively. COS-CVAE~\cite{mahajan2020diverse} aims to learn \emph{context-object split} factorization of the latent variables for an image. Different from these work, we model the disentanglement on time series medical images. Besides, we provide an identifiability result together with a reformulated sequential variational auto-encoder, enable the learning of the disease-causative hidden variable (without mixing others).

\section{Preliminaries}

We provide a brief background of structural causal model (SCM) and one can refer to \cite{pearl2009causality} for more details. 

\noindent \textbf{Structural Causal Model.} The SCM, according to \cite{pearl2009causality}, refers to a causal graph associated with the structural equations. The causal graph is represented by a directed acylic graph (DAG) denoted as $G := (V,E)$ with $V,E$ respectively denoting the node set and the edge set. Each arrow $x \to y$ in the $E$ denotes that the $x$ has a direct effect on $y$, \textit{i.e.}, fixing other nodes in $V$ except $x,y$, changing the value of $x$ would change the distribution of $y$. The structural equations assign the generating mechanisms of each variable in $V$. Specifically, for $V:=\{v_1,...,v_k\}$, the causal mechanisms associated with the structural equations (defined as $\{f_i\}_{v_i \in V}$) are defined as: $\{v_i \gets f_i(Pa(v_i),\varepsilon_i)\}_{v_i \in V}$ with $Pa(v_i)$ denoting the set of parent nodes of $v_i$. The $\{ \varepsilon_i \}_{v_i \in V}$ denotes the exogenous variables (\textit{i.e.}, the ones are not of interest/outside the system of the causal graph), which induce the distribution of $p(v_i|Pa(v_i))$. According to Causal Markov Condition \cite{pearl2009causality}, we have that $p(v_1,...,v_k) = \Pi_i p(v_i|Pa(v_i))$. The SCM can enable the definition of confounding bias which can induce correlation rather than causation, such as the correlation between $v_1$ and $v_3$, due to confounder $v_2$ in $v_1 \gets v_2 \to v_3$.

\section{Methodology}



\noindent \textbf{Problem Setup $\&$ Notations.} Denote $\bm{x}_t \in \mathcal{X}_t,y_t \in \mathcal{Y},\bm{A}_t \in \mathcal{A}, \bm{B}_{t} \in \mathcal{B}$ respectively as the image, disease status, clinical measurements and personal attributes at time stage $t$. Here the $\mathcal{Y} := \{\pm 1\}$ with $+1$ denoting the disease and $-1$ denoting the healthy status. We consider the disease progression problem, \textit{i.e.}, $p(y_T|\bm{u}_{t_1:t_2})$ with $\bm{u}:=\{\bm{x},\bm{A},\bm{B}\}$ and $t_{1} \leq t_2 < T$. To achieve this goal, we observe training data $\{\bm{u}^i_{t_1:t_2},y_T^i\}_{i \in [n]}$ with $[n] := \{1,...,n\}$. Note that we do not require observing $y_t$ for $t < T$, which agrees with many realistic scenarios in which the labelling can be costly.

\noindent \textbf{Outline.} We first introduce our causal hidden Markov Model in section~\ref{sec:model} in which only a subset of hidden variables are causally related to the disease (\textit{i.e.}, disease-causative features). Then we introduce our method to learn these disease-causative hidden variables in section~\ref{sec:learn}. Finally, we in section~\ref{sec:disentangle} provide an \emph{identifiability} claim which ensures that such disease-causative features can be disentangled with others. 

\begin{figure*}[tb]
\centering
\includegraphics[width=0.95\linewidth]{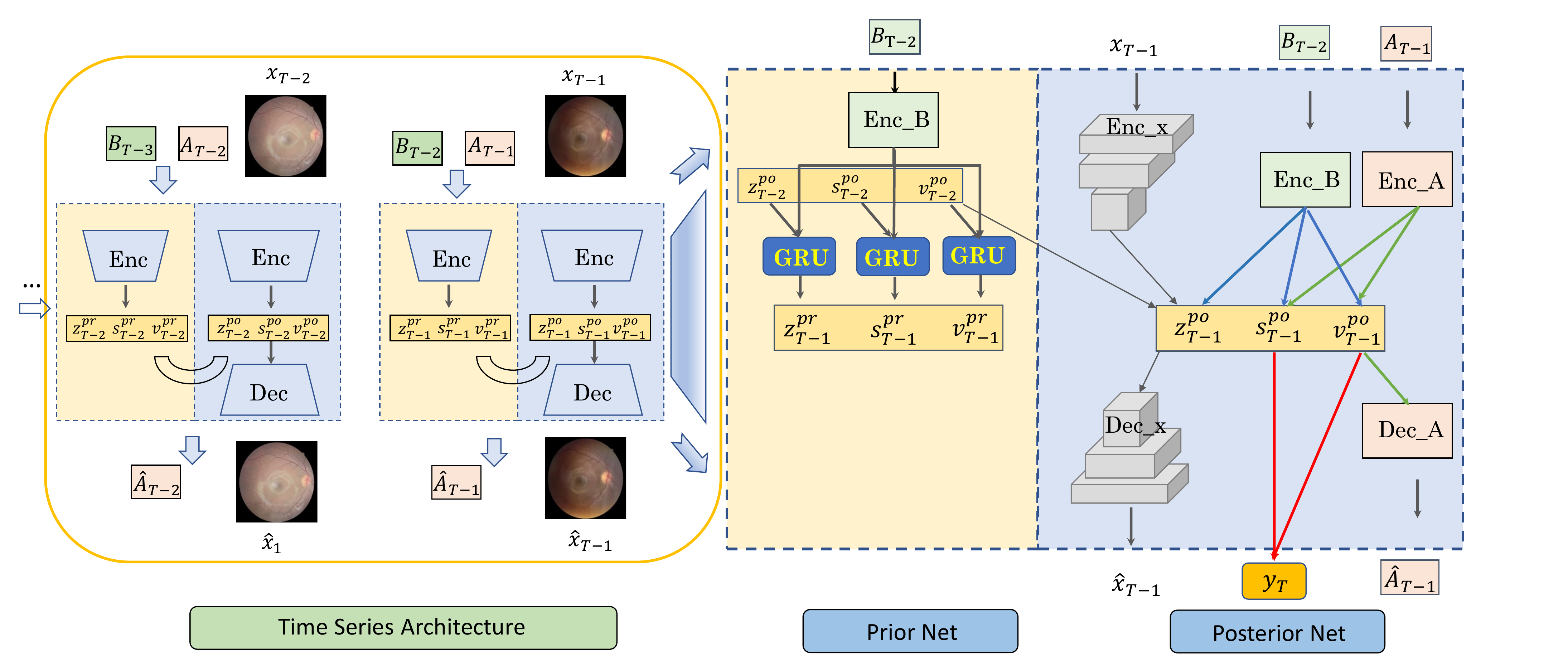}
\caption{
Left: illustration of the time series architecture for our proposed Causal-HMM. Right: the prior, posterior networks of the variational autoencoder at each time step. At each time step $t$, for the prior network, it takes the concatenated features (i. features via encoder on personal attributes $\bm{B}_{t-1}$; ii. the hidden variable at the last step ($\bm{z}_{t-1},\bm{s}_{t-1},\bm{v}_{t-1}$) into GRU. 
The GRU is followed with two FC layers which output the mean and log-variance vectors respectively. After sampling, the prior hidden variables ($\bm{z}_{t}^{pr},\bm{s}_{t}^{pr},\bm{v}_{t}^{pr}$) are obtained.
For the posterior network, the image $\bm{x}_t$ is first processed by five-layer convolution to extract image features, meanwhile the $\bm{A}_t$ and $\bm{B}_{t-1}$ are processed by fully connected layers to extract corresponding attribute features. The features are processed by two fully connected layers which output the mean and log-variance vectors for the posterior hidden variables ($\bm{z}_{t}^{po},\bm{s}_{t}^{po},\bm{v}_{t}^{po}$).
The hidden variables ($\bm{z}_{t}^{po},\bm{s}_{t}^{po},\bm{v}_{t}^{po}$) are then fed into decoder network for reconstruction of $\bm{x}_t$ and $\bm{A}_t$. Finally at time step $T-1$, the $\bm{s}_{T-1}^{po},\bm{v}_{T-1}^{po}$ are fed into the classifier to predict the future disease label $y_T$.
} 
\label{fig:network}
\end{figure*} 

\subsection{Causal Hidden Markov Model}
\label{sec:model}

To describe the disease progression, we introduce our causal graph with the DAG illustrated in Fig.~\ref{fig:framework}. As an extension of hidden Markov model to the supervised learning with consideration of disease-causative features, our model is named as \textbf{Causal} \textbf{H}idden \textbf{M}arkov \textbf{M}odel (Causal-HMM), which is formally defined as:
\begin{definition}[Causal-HMM]
\label{def:CHMM}
The structural equations $\mathcal{F}$ following the framework of Structural Causal Models \cite{pearl2009causality} of our Causal-HMM associated with the DAG in Fig.~\ref{fig:framework} is defined as $\mathcal{F}:=\{\mathcal{F}_t\}_t$ with $\mathcal{F}_t := \{\bm{B}_{t-1} \gets f_B(\varepsilon^t_b),\bm{v}_t \gets f_v(\bm{B}_{t-1},\varepsilon^t_v), \bm{s}_t \gets f_s(\bm{B}_{t-1},\varepsilon^t_s), \bm{z}_t \gets f_z(\bm{B}_{t-1},\varepsilon^t_z), \bm{A}_t \gets f_A(\bm{v}_t,\varepsilon^t_A), \bm{x}_t \gets f_x(\bm{s}_t,\bm{v}_t,\bm{z}_t,\varepsilon^t_x)\}$ for $t<T$ and $\mathcal{F}_T$ additionally contains $y_T \gets$ $f_Y(\bm{s}_t,\bm{v}_t,\varepsilon^T_y)$. $\{\{\varepsilon^t_B,\varepsilon^t_s,\varepsilon^t_v,\varepsilon^t_z,\varepsilon^t_x,\varepsilon^t_A\}_{t<T},\varepsilon^T_y\}$ are independent exogenous variables. The $\{\varepsilon^t_{B}\}_{t\leq T}$ (also the $\{\varepsilon^t_{s}\}_{t\leq T},\{\varepsilon^t_{v}\}_{t\leq T},\{\varepsilon^t_{z}\}_{t\leq T}$, $\{\varepsilon^t_{x}\}_{t\leq T},\{\varepsilon^t_{A}\}_{t\leq T}$) are same distributed with respect to $t$.
\end{definition}
\vspace{-1mm}
To have an intuitive understanding regarding our model, as shown in Fig.~\ref{fig:framework}, we introduce hidden variables $\bm{s}_t,\bm{v}_t,\bm{z}_t$ to model the latent components of observed variables $\bm{x}_t,\bm{A}_t,y_t$ at time step $t$. These hidden variables, which evolve as the intrinsic drive of the progression of the image $\bm{x}_t$ and the disease status, are additionally affected by an auxiliary variable $\bm{B}_{t-1}$ (\textit{i.e.}, the personal attributes such as age, gender that characterize the population), as reflected by the arrow $\bm{B}_{t-1} \to \bm{s}_t, \bm{v}_t, \bm{z}_t$ in Fig.~\ref{fig:framework}. Such an auxiliary variable $\bm{B}_t$ can explain the distributional change among populations. Besides, we explicitly separate the hidden variables into three parts: $\bm{s}_t,\bm{v}_t,\bm{z}_t$ that participant in different generating process. For disease forecasting, the $\bm{v}_t$ refer to components determining the clinical measurements $A_t$ related to the lesion region of the disease; the $\bm{s}_t$ denotes additional disease-causative factors that beyond properties but can be reflected in the image; the $\bm{z}_t$ denotes other concepts that are outside (but can be correlated to) the lesion region. In other words, all these latent variables generate $\bm{x}_t$; but among them, only $\bm{s}_t,\bm{v}_t$ point to $y_t$ with $\bm{v}_t$ additionally pointing to $\bm{A}_t$. Such a disentanglement of disease-causative hidden variables (\textit{i.e.}, $\bm{s}_t,\bm{v}_t$) from others, is the key to avoid spurious correlation. In the subsequent section, we provide the learning method for the proposed Causal-HMM and our identifiability result which implicitly ensures this disentanglement during learning.

\subsection{Learning Method}
\label{sec:learn}
To learn the proposed causal hidden Markov model, we introduce a new reformulated sequential VAE framework based on VAE, with the network architecture shown in Fig.~\ref{fig:network}.
The ELBO with $q_{\phi}(\bm{h}_{1:T-1}|\bm{u}_{1:T-1},y_T)$ ($\bm{h} := \{\bm{s},\bm{v},\bm{z}\}$,$\bm{u}:=\{\bm{x},\bm{A},\bm{B}\}$ for simplicity) as variational distribution is:
\begin{align}
    \label{eq:elbo-time}
    \mathbb{E}_{p(\bm{u}_{<T},y_T)} \left( \mathcal{L}_{q_{\phi},p_{\psi}} \right),
\end{align}
where $\mathcal{L}_{q_{\phi},p_{\psi}} =  \left[ \mathbb{E}_{q_{\phi}(\bm{h}_{<T}|\bm{u}_{<T},y_T)} \log{\left( \frac{p_{\psi}(\bm{h}_{<T},\bm{u}_{<T}, y_T)}{q_{\phi}(\bm{h}_{<T}|\bm{u}_{<T},y_T)} \right) } \right]$. According to Causal Markov Condition \cite{pearl2009causality}, we have the following factorization of joint distribution as:
\begin{align}
    \label{eq:fac-join}
    & p(\bm{h}_{<T},\bm{u}_{<T}, y_T) = p(y_T | \bm{s}_{T-1}, \bm{v}_{T-1}) * \nonumber \\
    & \Pi_{t=1}^{T-1} \Big( p(\bm{h}_t | \bm{h}_{t-1}, \bm{B}_{t-1})p(\bm{x}_t|\bm{h}_t)p(\bm{A}_t |\bm{v}_t) \Big).
\end{align}
The\quad$\{p(\bm{h}_t|\bm{h}_{t-1},\bm{B}_{t-1})\}_t$,$q_{\phi}(\bm{h}_{1:T-1}|\bm{u}_{1:T-1},y_T)$, $\{\{p(\bm{x}_t|\bm{h}_t),p(\bm{A}_t |\bm{v}_t)\}_t, p(y_T | \bm{s}_{T-1}, \bm{v}_{T-1}) \}$ are respectively prior models, posteriors models and generative models.
Note that the Markov property often exists on disease progression~\cite{Guihenneuc2000,Liu2013Markov} which we also adopt to our learning procedure.

\noindent \textbf{Prior.} For the prior $p_{\psi}(\bm{h}_t | \bm{h}_{t-1}, \bm{B}_{t-1})$, it can be further factorized due to disentanglement of $\bm{s},\bm{z},\bm{v}$: 
\begin{align}
\label{eq:prior}
p_{\psi}(\bm{h}_t | \bm{h}_{t-1}, \bm{B}_{t-1}) = \Pi_{\bm{o}} p_{\psi}(\bm{o}_t | \bm{o}_{t-1}, \bm{B}_{t-1}),
\end{align}
where for any $\bm{o} \in \{\bm{s}, \bm{v}, \bm{z}\}$, $p_{\psi}(\bm{o}_t | \bm{o}_{t-1}, \bm{B}_{t-1})$ for each $t$ is distributed as $\mathcal{N}(\mu_{\psi}(\bm{o}_{t-1}, \bm{B}_{t-1}), \Sigma_{\psi}(\bm{o}_{t-1}, \bm{B}_{t-1}))$. 
The $\{\mu_{\psi}(\bm{o}_{t-1}, \bm{B}_{t-1})\}_t$ (and $\{\Sigma_{\psi}(\bm{o}_{t-1}, \bm{B}_{t-1})\}_t$) parameterized by Gated Recurrent Unit (GRU) network \cite{chung2014empirical}.  The GRU is to capture the one-step dependency which has been employed in \cite{Zheng2015}; and following GRU is two FCs with one outputting the mean vector and one outputting the log-variance vector of the hidden variable.

\noindent \textbf{Posterior.} 
Since $q_{\phi}$ is expected to mimic the behavior of $p_{\psi}$ (also $p$), it shares the same way of reparameterization with $p_{\psi}=p_{\psi}(\bm{h}_{<T}|\bm{u}_{<T},y_T)$. Under reparameterization with $p_{\psi}$ and mean-field factorization\footnote{Please refer to supplementary for more details.}, the posterior is given by:
\begin{align}
    \label{eq:pos}
    q_{\phi}(\bm{h}_{<T}|\bm{u}_{<T},y_T) & = \frac{q_{\phi}(y_T|\bm{s}_{T-1},\bm{v}_{T-1})}{q_{\phi}(y_T|\bm{u}_{<T})} \nonumber \\
    & \quad *\Pi_{t<T} q_{\phi}(\bm{h}_{t}|\bm{u}_{t},\bm{h}_{t-1}),
\end{align} 
where $q_{\phi}(\bm{h}_{t}|\bm{u}_{t},\bm{h}_{t-1}) \sim \mathcal{N}(\mu(\bm{h}_{t-1},\bm{u}_{t}),\Sigma(\bm{h}_{t-1},\bm{u}_{t}))$.

Specifically, the posterior network $q_{\phi}(\bm{h}_t|\bm{h}_{t-1},\bm{u}_t)$ is parameterized by a five-layer convolution for encoding image and two fully connected layers for encoding the clinical measurements $\bm{A}$ and personal attributes $\bm{B}$ respectively. 

\noindent \textbf{Generations.} For each $t$, the generative models $p_{\psi}(\bm{x}_t|\bm{h}_t)$, $p_\phi(\bm{A}_t|v_t)$ are p.d.f of Gaussian distributions respectively parameterized by composition of deconvolution and that of fully connected (fc) layer, to reconstruct the image and the clinical measurements. The $q_{\psi}(y_T|\bm{s}_{T-1},\bm{v}_{T-1})$ is parameterized as an fc layer followed by softmax classifier.  The generation model for image $p_{\psi}(\bm{x}_t|\bm{h}_t)$ is parameterized by five-layer deconvolution;

\noindent \textbf{Reformulation.} Substituting the posterior in Eq.~\eqref{eq:pos} and the prior in Eq.~\eqref{eq:prior} in Eq.~\eqref{eq:elbo-time}, we reformulate the ELBO as: 
\begin{align}
    & \mathbb{E}_{p(\bm{u}_{<T},y_T)}\left[ \log{q_{\phi}(y_T|\bm{u}_{<T})} + 
    \sum_{t=1}^{T-1} \mathcal{L}^t_{q_\phi,p_\psi} \right] \label{eq:re-elbo} \\
    & \mathcal{L}^t_{q_\phi,p_\psi} = \mathbb{E}_{q_{\phi}(\bm{h}_{t}|\bm{u}_{t},\bm{h}_{t-1})} \left[ 
     \log{(p_{\psi}(\bm{x}_t|\bm{h}_t)*p_{\psi}(\bm{A}_t|\bm{v}_t)})  \right] \nonumber \\
    & \qquad  -D_{\mathrm{KL}}(q_{\phi}(\bm{h}_{t}|\bm{u}_{t},\bm{h}_{t-1}),p_{\psi}(\bm{h}_t | \bm{h}_{t-1}, \bm{B}_{t-1}))  \label{eq:l-t} \\
    & \mathcal{L}^{T-1}_{q_\phi,p_\psi} = \mathbb{E}_{q_{\phi}(\bm{h}_{T-1}|\bm{u}_{T-1},\bm{h}_{T-2})}\left[ (\ell_1 + \ell_2 + \ell_3) \right],
\end{align}
where the $\ell_1,\ell_2,\ell_3$ are respectively defined as:
\begin{align*}
\ell_1 & := \log{\left( p_{\psi}(\bm{x}_{T-1}|\bm{h}_{T-1})*p_{\psi}(\bm{A}_{T-1}|\bm{v}_{T-1}) \right)} \\
\ell_2 & :=  \log{\left( \frac{p_{\psi}(y_T|\bm{s}_{T-1},\bm{v}_{T-1})}{q_{\phi}(y_T|\bm{s}_{T-1},\bm{v}_{T-1})} \right)} \\
\ell_3 & := \log{\frac{p_{\psi}(\bm{h}_{T-1} | \bm{h}_{T-2}, \bm{B}_{T-2})}{q_{\phi}(\bm{h}_{T-1}|\bm{u}_{T-1})}}.  
\end{align*}
Due to the approximation of $p_{\psi}$ by $q_{\phi}$, we parameterize the $p_{\psi}(y_T|\bm{s}_{T-1},\bm{v}_{T-1})$ as $q_{\phi}(y_T|\bm{s}_{T-1},\bm{v}_{T-1})$, with which the $\ell2$ degenerates to 0. Besides, we have for $q_\phi(y_T|\bm{u}_{<T})$:
\begin{align*}
    \int \left(\Pi_{t=1}^{T-1} q_\phi(\bm{h}_t|\bm{u}_t,\bm{h}_{t-1}) \right) q_{\phi}(y_T|\bm{s}_{T-1},\bm{v}_{T-1}) d\bm{h}_0...d\bm{h}_{T-1}.
\end{align*}

\noindent \textbf{Training \& Test.} With such reparameterizations, the reformulated ELBO in Eq.~\eqref{eq:re-elbo} is our maximization objective. During the inference, we iteratively obtain the latent variable $\bm{h}_t$ at each step via the posterior network $q_\phi(\bm{h}_t|\bm{u}_t,\bm{h}_{t-1})$. Finally, we feed $\bm{s}_{T-1},\bm{v}_{T-1}$ into the the classifier $q_{\phi}(y_T|\bm{s}_{T-1},\bm{v}_{T-1})$ to predict $y_T$.

\begin{table*}
\begin{center}
\resizebox{\linewidth}{!}{
\small
\begin{tabular}{|c |c c |c c| c c|c c|}
\hline
Methods
&\multicolumn{2}{c|}{\bf \textbf{RGL}~\cite{louis2019riemannian}}
&\multicolumn{2}{c|}{\bf \textbf{Devised RNN} \cite{Gao2018BrainDD}} 
&\multicolumn{2}{c|}{\bf \textbf{LogSparse Transformer} \cite{li2019enhancing}} 
&\multicolumn{2}{c|}{\bf Ours}\\ 
\hline
Grades & ACC & AUC & ACC & AUC & ACC & AUC  & ACC & AUC \\
\hline
G1 to G5       & 64.70 $\pm$ 1.89 &  71.14 $\pm$ 1.44  &  74.02 $\pm$ 3.82 & 81.75 $\pm$ 3.08 &   74.58 $\pm$ 4.81  & 79.2 $\pm$ 4.69 & \textbf{77.19} $\pm$ 1.69  & \textbf{85.43} $\pm$ 1.76 \\
\hline
G1 to G4       & 61.46 $\pm$ 4.48 &  63.50 $\pm$ 5.88 & 66.92 $\pm$ 1.06 & 72.88 $\pm$ 2.22 &  70.42 $\pm$ 4.57  & 73.95 $\pm$ 3.98   &  \textbf{72.89} $\pm$ 2.64  & \textbf{78.99} $\pm$ 1.53 \\
\hline
G1 to G3     & 58.33 $\pm$ 10.31 &  56.64 $\pm$ 5.41   & 63.55 $\pm$ 1.47 & 66.45 $\pm$ 0.92 &  \textbf{67.18} $\pm$ 1.99  & \textbf{70.15} $\pm$ 1.1 &  62.43 $\pm$ 2.03  & 68.24 $\pm$ 2.93 \\
\hline
G1 to G2  & 64.17 $\pm$ 2.03 &  53.38 $\pm$ 4.20   &  57.19 $\pm$ 7.45 & 57.07 $\pm$ 2.37  &  63.13 $\pm$ 4.01  & \textbf{65.15} $\pm$ 2.12 &  \textbf{65.42} $\pm$ 1.47  & 65.09 $\pm$ 2.29\\
\hline
G2 to G5    & 62.29 $\pm$ 5.63 &  69.70 $\pm$ 5.12   &   73.27 $\pm$ 2.44 & 80.16 $\pm$ 1.48 &  76.04 $\pm$ 0.74  & 84.02 $\pm$ 1.76 &  \textbf{76.26} $\pm$ 2.44  & \textbf{86.71} $\pm$ 0.89 \\
\hline
G2 to G4    & 61.25 $\pm$ 7.19 &  65.08 $\pm$ 5.68 &   67.10 $\pm$ 1.79 & 74.21 $\pm$ 1.59  &  \textbf{72.71} $\pm$ 2.13  & 80.03 $\pm$ 2.19  &  71.22 $\pm$ 5.17  & \textbf{80.62} $\pm$ 1.36\\
\hline
G2 to G3   & 58.12 $\pm$ 5.68 & 56.40 $\pm$ 4.98 &  63.17 $\pm$ 1.94 & 66.87 $\pm$ 2.92  & 65.62 $\pm$ 4.54  & 67.79 $\pm$ 4.49  &  \textbf{66.91} $\pm$ 2.69  & \textbf{75.07} $\pm$ 1.31 \\
\hline
G3 to G5  & 65.62 $\pm$ 4.60 &  71.22 $\pm$ 5.50    & 74.77 $\pm$ 3.03 & 80.57 $\pm$ 2.28 &  76.45 $\pm$ 4.81  & 83.16 $\pm$ 3.56 &  \textbf{77.01} $\pm$ 3.41  & \textbf{86.22} $\pm$ 1.34 \\
\hline
G3 to G4    & 63.54 $\pm$ 1.95 &  67.58 $\pm$ 3.02& 68.79 $\pm$ 4.46 & 73.49 $\pm$ 3.19  &  \textbf{72.49} $\pm$ 4.81  & 77.41 $\pm$ 1.99 &  71.77 $\pm$ 2.59  & \textbf{82.22} $\pm$ 1.29 \\
\hline
G4 to G5  & 67.29 $\pm$ 3.42 &  71.47 $\pm$ 3.77   &  75.53 $\pm$ 3.07 & 81.81 $\pm$ 2.12 &  \textbf{79.58} $\pm$ 2.16  & 86.69 $\pm$ 1.52 &  78.13 $\pm$ 3.21  & \textbf{86.92} $\pm$ 1.53  \\
\hline
Mean   & 62.68 $\pm$ 4.72 &  64.41 $\pm$ 4.50   & 68.43 $\pm$ 3.05 & 73.53 $\pm$ 2.22 &  71.82 $\pm$ 3.46  & 76.76 $\pm$ 2.74 &  \textbf{71.92} $\pm$ 2.73  & \textbf{79.55} $\pm$ 1.53 \\
\hline
\end{tabular}
}
\end{center}
\caption{Comparison results over other methods. Results of ACC (accuracy, mean $\pm$ std $\%$) and AUC (Area Under the Curve, mean $\pm$ std $\%$) on the test dataset between ours with \textbf{RGL}~\cite{louis2019riemannian}, \textbf{Devised RNN}~\cite{Gao2018BrainDD} and \textbf{LogSparse Transformer}~\cite{li2019enhancing} on 10 time series settings.}
\label{tab:comparison}
\end{table*}

\subsection{Identifiability of Disease-Causative Features}
\label{sec:disentangle}

In this section, we provide a theoretical guarantee for our learning method that the disease-causative features at each $t$ (\textit{a.k.a.} $\bm{s}_t,\bm{v}_t$) \cite{sun2020latent}, can be disentangled from others that may encode spurious correlations (\textit{a.k.a.} $\bm{z}_t$), ensuring the stable learning of our method. Our analysis is inspired but far beyond the recent result \cite{khemakhem2019variational} in nonlinear ICA to the supervised learning with time-series graphical model, in which the main objective is to disentangle the $\bm{s}_t,\bm{v}_t$ from $\bm{z}_t$ at each time step $t$ in order to avoid spurious correlation. Similar to \cite{khemakhem2019variational,khemakhem2020ice}, we assume the $\bm{x},\bm{A},y$ are generated by \textit{Additive Noise Model} (ANM), which can be a wide class of continuous and categorical distributions \cite{janzing2009identifying}. Besides, we assume that the latent variables $p(\bm{s}_t,\bm{v}_t,\bm{z}_t|\bm{B}_{j\leq t-1})$ for every $t \in [T]$) belong to the exponential family:
\begin{align}
    & p_{\bm{T}^t,\bm{\Gamma}^t}(\bm{s}_t,\bm{v}_t,\bm{z}_t|\bm{B}_{t-1}) = \Pi_{\bm{o} \in \{\bm{s},\bm{v},\bm{z}\}} p_{\bm{T}^t_{\bm{o}},\bm{\Gamma}^t_{\bm{o}}}(\bm{o}_t|\bm{B}_{ t-1}), \nonumber \\
    & p_{\bm{T}^t_{\bm{o}},\bm{\Gamma}^t_{\bm{o}}}(\bm{o}_t|\bm{B}_{t-1}) = \nonumber \\ 
    & {\small\prod_{i=1}^{d_o}} \frac{C^t_i(o_i)}{Q^t_{\bm{o},i}} \exp\Big( {\small \sum_{k=1}^{k_o}} T^t_{\bm{o},i,k}(o_i) \Gamma^{t}_{\bm{o},i,k}(\bm{B}_{t-1})  \Big) \nonumber
\end{align}
 for any $\bm{o} \in \{\bm{s}, \bm{v}, \bm{z}\}$. Here the $\{T^t_{\bm{o},i,k}(o_i)\}, \{\Gamma^t_{\bm{o},i,k}\}$ denote the sufficient statistics and natural parameters; and the $\{C^t_i\}, \{Q^t_{\bm{o},i}\}$ denote the base measures and normalizing constants to ensure the integral of distribution equals to 1. Let $\mathbf{T}^t_{\bm{o}} \!:=\! \left[\mathbf{T}^t_{\bm{o},1},...,\mathbf{T}^t_{\bm{o},d_o}  \right]$ $\!\in\! \mathbb{R}^{k_o \times d_o}$ $\big(\mathbf{T}^t_{\bm{o},i} \!:=\! [T^t_{\bm{o},i,1},...,T^t_{\bm{o},i,k_o}], \forall i \in [d_o]\big)$ and $\bm{\Gamma}^t_{\bm{o}} \!:=\! \left[\Gamma^t_{\bm{o},1},...,\Gamma^t_{\bm{o},d_o} \right]$ $\!\in\! \mathbb{R}^{k_o \times d_o}$ $\big(\bm{\Gamma}^t_{\bm{o},i} \!:=\! [\Gamma^t_{\bm{o},i,1},...,\Gamma^t_{\bm{o},i,k_o}], \forall i \in [d_o]\big)$. Then we have the following identifiability result for $\theta := \{\{\mathbf{T}^{t\leq T}_{\bm{o}}\}_{\bm{o}},\{\bm{\Gamma}^{t\leq T}_{\bm{o}}\}_{\bm{o}},f_x,f_y,f_A\}$:

\begin{theorem}[Identifiability]
\label{thm:iden}
We assume that $f_{x},f_{y},f_A$ are bijective. Denote $g^t_{y}(s) := \mathbb{E}(y_T|s_t,v_t,\bm{B}_{j\leq t})$. Under the following conditions:
\begin{itemize}[topsep=0pt,itemsep=-1ex]
    \item $\{T^t_{\bm{o},i,j}\}$ are differentiable and non-zero almost everywhere for any $\bm{o} \in \{\bm{s},\bm{v},\bm{z}\}$ and $t \leq T$.
    \item For every $t$, there exists at least $m := d*k+1$ with $d:=$ $\max(d_s,d_v,d_z)$ and $k:=\max(k_s,k_v,k_z)$ values of $\bm{B}_{t=0}$, \textit{i.e.}, $\bm{B}_{1,t=0},...,\bm{B}_{m,t=0}$ such that the 
        $[\bm{\Gamma}^t_{\bm{o}}(\bm{B}_{2,t=0})- \bm{\Gamma}^t_{\bm{o}}(\bm{B}_{1,t=0}),...,$ $\bm{\Gamma}^t_{\bm{o}}(\bm{B}_{m,t=0})- \bm{\Gamma}^t_{\bm{o}}(\bm{B}_{1,t=0})]$
        have full column rank and $\bm{o} \in \{\bm{s},\bm{v},\bm{z}\}$,
\end{itemize}
we have that if $\theta$ and $\tilde{\theta}$ give rise to the same observational distribution, \textit{i.e.}, $p_\theta(\bm{x}_t, y_T,\bm{v}_t) = p_{\tilde{\theta}}(\bm{x}_t,y_T,\bm{v}_t)$ for any $\bm{x}_t,y_T,\bm{v}_t$ and $t < T$, then there exists invertible matrices $\{M^t_{\bm{o}}\}_{\bm{o} \in \{ \bm{s},\bm{v},\bm{z} \}}$ and vectors $\{b^t_{\bm{o}}\}_{\bm{o} \in \{ \bm{s},\bm{v},\bm{z} \}}$ such that: 
\begingroup
\allowdisplaybreaks
\begin{align}
& \text{Disentangle}:  \nonumber \\ 
& \qquad T^t_{\bm{s}}([f^{-1}_{x}]_{\mathcal{S}}(x_t)) = M^t_{\bm{s}} \tilde{T}^t_{\bm{s}}([\tilde{f}^{-1}_{x}]_{\mathcal{S}}(x_t)) + b^t_{\bm{s}}, \label{eq:iden-s}\\
& \qquad T^t_{\bm{v}}([f^{-1}_{x}]_{\mathcal{V}}(x_t)) =  M^t_{\bm{v}} \tilde{T}^t_{\bm{v}}([\tilde{f}^{-1}_{x}]_{\mathcal{V}}(x_t)) + b^t_{\bm{v}}, \label{eq:iden-a}\\
& \qquad T^t_{\bm{z}}([f^{-1}_{x}]_{\mathcal{Z}}(x_t))  = M^t_{\bm{z}} \tilde{T}^t_{\bm{z}}([\tilde{f}^{-1}_{x}]_{\mathcal{Z}}(x_t)) + b^t_{\bm{z}}, \label{eq:iden-z}\\
& \text{Prediction}:  \nonumber \\ 
& \qquad \tilde{g}^t_y([f^{-1}_{x}]_{\mathcal{S},\mathcal{V}}(x_t),\bm{B}_{t}) = \tilde{g}^t_y([\tilde{f}^{-1}_{x}]_{\mathcal{S},\mathcal{V}}(x_t),\bm{B}_t) \label{eq:iden-y}.
\end{align}
\endgroup
\end{theorem}
\begin{remark}
The $g_y^t$ is related to $f_s, \bm{B}_{j \geq t}, f_y$. The second ``full-column" rank condition, as an indication of independence of natural parameters $\bm{\Gamma}$, implies that the distributions with different personal attributes (population) are diverse enough, which is also assumed in \cite{khemakhem2019variational}.
\end{remark}

\vspace{-0.2mm}
Note that the Eq.~\eqref{eq:iden-s},~\eqref{eq:iden-a},~\eqref{eq:iden-z} imply the disentanglement of $\bm{s},\bm{z},\bm{v}$ unless the extreme case that these three latent components can be represented by each other, \textit{i.e.}, there exists $h$ such that $h([f^{-1}(x)]_{\mathcal{S}}) = [f^{-1}(x)]_{\mathcal{V}}$ or $h([f^{-1}(x)]_{\mathcal{Z}}) = [f^{-1}(x)]_{\mathcal{V}}$. Besides, the Eq.~\eqref{eq:iden-y} shows that we can learn the same prediction at time $t$ to the ground-truth (consider the $\theta:=\theta^\star$ as the ground-truth oracle parameter and the $\tilde{\theta}$ denotes or learned parameter), under the deterministic setting (\textit{i.e.}, $\varepsilon_x = 0$). Such an identifiability result ensures the disentanglement of our method. Besides, it does not contradict with the conclusion in \cite{locatello2019challenging} of ``impossibility to learn disentangled representations without supervision", since our learning is additionally supervised by the label $y$, the clinical measurements $\bm{A}$ and also guided by the personal attributes $\bm{B}$ as side information.

\section{Experiments and Analysis}

In this section, we apply our method on in-house data that studies the peripapillary atrophy (PPA) development among primary school students. The atrophy happens around the region of the optic disc (as marked by the red rectangle in Fig.~\ref{fig:visualization}). Since the PPA can cause irreversible myopia retinas in children, the forecast of it at an early stage is extremely valuable to slow down the progression.

\subsection{Dataset}

\begin{table*}
\begin{center}
\resizebox{\linewidth}{!}{
\small
\begin{tabular}{|c |c c |c c| c c |c c | c c|c c|}
\hline
Methods
&\multicolumn{2}{c|}{\bf CNN}
&\multicolumn{2}{c|}{\bf CNN+LSTM}
&\multicolumn{2}{c|}{\bf Seq VAE}
&\multicolumn{2}{c|}{\bf Seq VAE + Att}
&\multicolumn{2}{c|}{\bf Ours}
\\ 
\hline
Grades & ACC & AUC & ACC & AUC & ACC & AUC & ACC & AUC & ACC & AUC \\
\hline
G1 to G5      &  62.17 $\pm$ 3.97  & 65.01 $\pm$ 1.48  & 74.39 $\pm$ 1.56 & 80.25 $\pm$1.93 & 75.40 $\pm$ 1.08  & 81.67 $\pm$ 1.66 & 74.21 $\pm$ 4.79 &  84.46 $\pm$ 1.54 & 77.19 $\pm$ 1.69  & 85.43 $\pm$ 1.76 \\
\hline
G1 to G4      &  61.64 $\pm$ 1.04  & 60.99 $\pm$ 2.42  & 69.36 $\pm$ 2.91 & 72.39 $\pm$ 3.16 & 68.78 $\pm$ 4.10  & 73.50 $\pm$ 1.99  & 71.21 $\pm$ 2.75 &  76.68 $\pm$ 2.08 & 72.89 $\pm$ 2.64  & 78.99 $\pm$ 1.53 \\
\hline
G1 to G3   &  58.07 $\pm$ 3.75  & 56.75 $\pm$ 1.99  & 61.31 $\pm$ 1.42 & 65.48 $\pm$ 2.23  & 60.75 $\pm$ 1.93  & 67.05 $\pm$ 2.75  & 60.78 $\pm$ 0.93 &  61.34 $\pm$ 0.91 &  62.43 $\pm$ 2.03  & 68.24 $\pm$ 2.93 \\
\hline
G1 to G2  &  59.25 $\pm$ 1.56  & 55.74 $\pm$ 1.76  & 59.44 $\pm$ 3.94 & 58.45 $\pm$ 3.21 & 62.62 $\pm$ 0.00  & 58.00 $\pm$ 1.62 & 62.43 $\pm$ 0.42 &  59.56 $\pm$ 4.79 &  65.42 $\pm$ 1.47  & 65.09 $\pm$ 2.29 \\
\hline
G2 to G5  &  64.11 $\pm$ 2.83  & 66.65 $\pm$ 1.14  & 70.84 $\pm$ 1.92 & 79.09 $\pm$ 1.15  & 72.52 $\pm$ 1.08  & 81.16 $\pm$ 0.91   & 77.19 $\pm$ 5.14 &  84.88 $\pm$ 1.17  &  76.26 $\pm$ 2.44  & 86.71 $\pm$ 0.89 \\
\hline
G2 to G4   &  61.23 $\pm$ 4.01  & 64.67 $\pm$ 1.25  & 69.34 $\pm$ 1.67 & 72.68 $\pm$ 1.56 & 68.85 $\pm$ 1.32  & 73.69 $\pm$ 1.75 & 74.39 $\pm$ 3.14 &  77.66 $\pm$ 0.99 &  71.22 $\pm$ 5.17  & 80.62 $\pm$ 1.36 \\
\hline
G2 to G3  &  59.25 $\pm$ 1.97  & 62.82 $\pm$ 2.07  & 61.49 $\pm$ 1.92 & 66.67 $\pm$ 2.74  & 64.02 $\pm$ 3.77  & 61.91 $\pm$ 3.08   & 61.31 $\pm$ 0.84 &  62.63 $\pm$ 3.46 &  66.91 $\pm$ 2.69  & 75.07 $\pm$ 1.31 \\
\hline
G3 to G5  &  67.35 $\pm$ 2.75  & 71.20 $\pm$ 1.61  & 74.21 $\pm$2.52 & 80.35 $\pm$ 1.28  & 74.53 $\pm$ 1.06  & 80.71 $\pm$ 1.96  & 76.63 $\pm$ 3.43 &  84.42 $\pm$ 1.80  &  77.01 $\pm$ 3.41  & 86.22 $\pm$ 1.34 \\
\hline
G3 to G4  &  62.89 $\pm$ 3.02  & 68.25 $\pm$ 2.98  & 68.59 $\pm$ 4.11 & 71.96 $\pm$ 6.56  & 67.29 $\pm$ 2.29  & 74.58 $\pm$ 0.85  & 74.77 $\pm$ 0.93 &  79.24 $\pm$ 0.99 &  71.77 $\pm$ 2.59  & 82.22 $\pm$ 1.29 \\
\hline
G4 to G5   &  70.75 $\pm$ 3.96  & 77.58 $\pm$ 1.37  & 75.51 $\pm$ 3.93 & 82.42 $\pm$ 2.51  & 73.83 $\pm$ 3.58  & 79.65 $\pm$ 1.65   & 76.45 $\pm$ 2.91 &  84.45 $\pm$ 0.48  &  78.13 $\pm$ 3.21  & 86.92 $\pm$ 1.53 \\
\hline
Mean  &  62.67 $\pm$ 2.89  & 64.97 $\pm$ 1.81  & 68.45 $\pm$ 2.59 & 72.97 $\pm$ 2.63 & 68.86 $\pm$ 2.02  & 73.19 $\pm$ 1.82 & 70.94 $\pm$ 2.53 &  75.53 $\pm$ 1.82 &  71.92 $\pm$ 2.73  & 79.55 $\pm$ 1.62 \\
\hline
\end{tabular}
}
\end{center}
\caption{Ablative study on our time series architecture, the attributes and the disentanglement. Results of ACC (accuracy, mean $\pm$ std $\%$) and AUC (Area Under the Curve, mean $\pm$ std $\%$) on the test dataset on 10 time series settings.}
\label{tab:ablation}
\end{table*}

\begin{table*}[ht]
\begin{center}
\resizebox{\linewidth}{!}{
\small
\begin{tabular}{|c |c c |c c|c c|c c |c c|c c|}
\hline
Variables
&\multicolumn{6}{c|}{\bf $\bm{s}$ + $\bm{v}$}
&\multicolumn{6}{c|}{\bf $\bm{z}$}\\
\hline
Dataset 
&\multicolumn{2}{c|}{\bf Training }
&\multicolumn{2}{c|}{\bf Validation }
&\multicolumn{2}{c|}{\bf Test }
&\multicolumn{2}{c|}{\bf Training }
&\multicolumn{2}{c|}{\bf Validation }
&\multicolumn{2}{c|}{\bf Test }
\\ 
\hline
Grades & ACC & AUC & ACC & AUC & ACC & AUC & ACC & AUC & ACC & AUC & ACC & AUC \\
\hline
G1 to G5     & 94.72 $\pm$ 1.96    & 99.40 $\pm$ 0.36   & 72.20 $\pm$ 1.10    & 75.10 $\pm$ 0.18  & 78.88 $\pm$ 1.06 & 83.75 $\pm$ 0.29 & 99.93 $\pm$ 0.16  & 100.00 $\pm$ 0.00 & 71.20 $\pm$ 1.64 & 78.10 $\pm$ 0.24 & 67.10 $\pm$ 1.22 & 68.98 $\pm$ 0.45  \\
\hline
G1 to G4      & 98.88 $\pm$ 1.81    & 99.86 $\pm$ 0.30   & 64.40 $\pm$ 1.34    & 66.06 $\pm$ 0.42  & 70.09 $\pm$ 0.66 & 70.55 $\pm$ 0.57 & 79.51 $\pm$ 6.79  & 87.94 $\pm$ 5.44 & 59.40 $\pm$ 3.29  & 60.96 $\pm$ 0.47 & 60.75 $\pm$ 1.14 & 63.32 $\pm$ 1.69\\
\hline
G1 to G3   & 79.51 $\pm$ 4.24    & 86.88 $\pm$ 3.76   & 59.20 $\pm$ 4.91    & 60.17 $\pm$ 0.55  & 59.81 $\pm$ 3.67 & 64.09 $\pm$ 0.87 & 58.68 $\pm$ 1.67 & 52.79 $\pm$ 3.89  & 63.00 $\pm$ 0.71 & 55.42 $\pm$ 1.48 & 61.68 $\pm$ 0.66 & 42.37 $\pm$ 0.66 \\
\hline
G1 to G2    & 69.51 $\pm$ 1.64    & 74.30 $\pm$ 1.25   & 70.80 $\pm$ 3.56    & 63.46 $\pm$ 0.64  & 69.16 $\pm$ 2.64 & 70.43 $\pm$ 0.47  & 79.65 $\pm$ 2.97  & 88.70 $\pm$ 3.28 & 65.40 $\pm$ 2.97 & 61.83 $\pm$ 0.53 & 53.83 $\pm$ 1.82 & 48.79 $\pm$ 1.41 \\
\hline
G2 to G5    & 98.13 $\pm$ 2.49    & 99.84 $\pm$ 0.33   & 74.40 $\pm$ 1.14    & 75.23 $\pm$ 0.09  & 78.13 $\pm$ 1.42 & 83.92 $\pm$ 0.21 & 100.00 $\pm$ 0.00  & 100.00 $\pm$ 0.00 & 70.60 $\pm$ 0.55 & 77.84 $\pm$ 0.47 & 66.92 $\pm$ 0.51 & 68.65 $\pm$ 0.20 \\
\hline
G2 to G4    & 97.78 $\pm$ 2.03    & 99.54 $\pm$ 0.81   & 63.40 $\pm$ 2.23    & 65.99 $\pm$ 0.38  & 70.47 $\pm$ 1.94 & 71.18 $\pm$ 0.91 & 85.83 $\pm$ 4.88  & 92.69 $\pm$ 4.29 & 58.60 $\pm$ 1.95 & 60.82 $\pm$ 0.97 & 60.75 $\pm$ 1.48 & 61.86 $\pm$ 0.56 \\
\hline
G2 to G3     & 77.15 $\pm$ 4.93    & 84.31 $\pm$ 5.19   & 64.40 $\pm$ 1.67    & 61.20 $\pm$ 0.78  & 64.30 $\pm$ 2.03 & 64.69 $\pm$ 2.31 & 56.81 $\pm$ 2.05  & 49.24 $\pm$ 0.59 & 63.00 $\pm$ 0.00 & 54.47 $\pm$ 1.71 & 62.63 $\pm$ 0.78 & 43.93 $\pm$ 2.06 \\
\hline
G3 to G5   & 96.81 $\pm$ 1.38    & 99.78 $\pm$ 0.16   & 74.00 $\pm$ 2.12    & 75.42 $\pm$ 0.37  & 77.94 $\pm$ 1.94 & 83.83 $\pm$ 0.30 & 100.00 $\pm$ 0.00  & 100.00 $\pm$ 0.00 & 70.60 $\pm$ 0.89 & 77.43 $\pm$ 0.28 & 66.17 $\pm$ 1.02 & 68.92 $\pm$ 0.23 \\
\hline
G3 to G4    & 99.10 $\pm$ 0.76    & 99.94 $\pm$ 0.08   & 64.40 $\pm$ 1.52    & 68.04 $\pm$ 0.82  & 68.04 $\pm$ 1.79 & 72.51 $\pm$ 0.51 & 85.49 $\pm$ 8.17  & 92.21 $\pm$ 5.74 & 60.80 $\pm$ 1.30 & 61.38 $\pm$ 0.98 & 60.75 $\pm$ 2.19 & 62.24 $\pm$ 2.19 \\
\hline
G4 to G5       & 95.07 $\pm$ 2.28    & 99.52 $\pm$ 0.42   & 72.20 $\pm$ 1.30    & 74.97 $\pm$ 0.18  & 78.13 $\pm$ 1.06 & 83.71 $\pm$ 0.11 & 99.38 $\pm$ 0.86  & 99.97 $\pm$ 0.05 & 70.40 $\pm$ 0.55 & 77.31 $\pm$ 0.34 & 67.66 $\pm$ 2.52 & 67.46 $\pm$ 0.46 \\
\hline
Mean  & 90.67 $\pm$ 2.35    & 94.34 $\pm$ 1.27   & 67.94 $\pm$ 2.09    & 68.56 $\pm$ 0.44  & \textbf{71.50} $\pm$ 1.82 & \textbf{74.87} $\pm$ 0.66 & 84.53 $\pm$ 2.76  & 86.35 $\pm$ 2.33 & 65.30 $\pm$ 1.39 & 66.56 $\pm$ 0.75 & \textbf{62.82} $\pm$ 1.33 & \textbf{59.65} $\pm$ 0.99 
\\
\hline
\end{tabular}
}
\end{center}
\caption{Ablative study on the different hidden variables of $\bm{s}+\bm{v}$ and $\bm{z}$ at the second stage. Results of ACC (accuracy, mean $\pm$ std $\%$) and AUC (Area Under the Curve, mean $\pm$ std $\%$) on training set, validation set and test set.
}
\label{tab:ablation_second_stage}
\end{table*}

Our data contains 507 sequential data, \textit{i.e.}, retinal images, clinical measurements and personal attributes of 507 students in primary school from the 1st grade to the 6th grade. Only the disease label at the 6th grade (\textit{i.e.}, $y_{T=6}$) are provided with $y_{t=1}$ being $-1$ (healthy) for all samples. 
The $\bm{x},\bm{A},\bm{B},y$ respectively denote the retinal image, clinical measurements related to the PPA, the personal attributes and the disease label. Our goal is to predict the disease label at the 6th grade (\textit{i.e.}, the $T=6$), given $\{\bm{x},\bm{A},\bm{B}\}_{t \in [t_1,t_2]}$ with $1 \leq t_1 < t_2 < T$. To validate the effectiveness of our method on handling OOD data, the sex ratio which has been found significantly correlated with disease progression \cite{czepita2019role, zhou2016five}, is different between the test data (boy/girl: 3/1) and the training, validation data (boy/girl: 2/3). After this splitting, the training, validation and test set contain 300, 100 and 107 samples, respectively. The clinical measurements $\bm{A}$ are represented by 15 vision-related attributes; while the personal attributes $\bm{B}$ are represented by 16 attributes \footnote{please refer to supplementary information for details.}.

\noindent \textbf{Gender v.s. Disease.} We conduct a Bilateral T-test of whether gender affects the disease. We assume the disease rate of boys and girls are respectively binomial distributions with parameters $p_{\mathrm{boy}},p_{\mathrm{girl}}$, denoted as $B(p_{\mathrm{boy}})$ and $B(p_{\mathrm{girl}})$.
The $H_0$ and $H_1$ hypothesis are:
$H_0: p_{\mathrm{boy}} = p_{\mathrm{girl}}, \ H_1: p_{\mathrm{boy}} \neq p_{\mathrm{girl}}$.
The calculated $p$-value in our dataset is 0.03876, implying that the probability of making an error if we admit $H_1$ (denying $H_0$) is less than $5\%$. Therefore, the datasets with different sex ratios can have the distributional difference, verifying that the distributions of training set and test set are different.

\subsection{Quantitative Results}
For comparison, we compare with the following methods. 1) \textbf{RGL}~\cite{louis2019riemannian} shares the most similar scenario with ours. They proposes to predict the disease progression by learning a Riemannian manifold space. To compare fairly, we additionally append a classifier after the latent space for disease prediction. 2) \textbf{Devised RNN}~\cite{Gao2018BrainDD} employs deep convolutional neural network and recurrent neural network to learn longitudinal features for disease classification. We also provide the attributes for its learning when adopting this method to our problem. 3) \textbf{LogSparse Transformer}~\cite{li2019enhancing} is a transformer-based method for time series forecasting. Similarly, to compare fairly, we additionally append a classifier after the transformer network. 

The prediction accuracy (ACC) and Area Under the ROC curve (AUC) are measured for evaluation. We consider $C_5^2=10$ time series settings for possible pairs of $t_1, t_2$ (\textit{e.g.}, the ``G1 to G5" means $t_1 = 1, t_2 = 5$). All results are shown in Tab.~\ref{tab:comparison}. As shown, our method achieves better result on both ACC and AUC than the compared baselines on almost all settings and the average setting. When comparing different settings inside one method, we see that using data more closer to the future stage leads to higher performance, showing that the closer stage contains more useful information for future disease. Note that looking further into the past does not boost the performance, we point that this is due to the current image has contained the sufficient information for future disease. What we should do on this sequential data is to prompt the future prediction by making use of the past time dependency. To compare in all settings, \textbf{Devised RNN}~\cite{Gao2018BrainDD} performs better than \textbf{RGL}~\cite{louis2019riemannian} benefiting from attribute data. The \textbf{LogSparse Transformer}~\cite{li2019enhancing} outperforms the first two due to the higher ability for learning time dependencies. While they are all weaker than ours. Especially ours outperforms the \textbf{RGL}~\cite{louis2019riemannian} by a large margin. Since such a latent space in \textbf{RGL} is for generating the whole image for disease prediction, it can mix the correlated by non-causative features of the disease. \textbf{Devised RNN}~\cite{Gao2018BrainDD} and \textbf{LogSparse Transformer}~\cite{li2019enhancing} also face the same problem. 

\begin{figure*}[!htbp]
\centering
\includegraphics[width=0.94\linewidth]{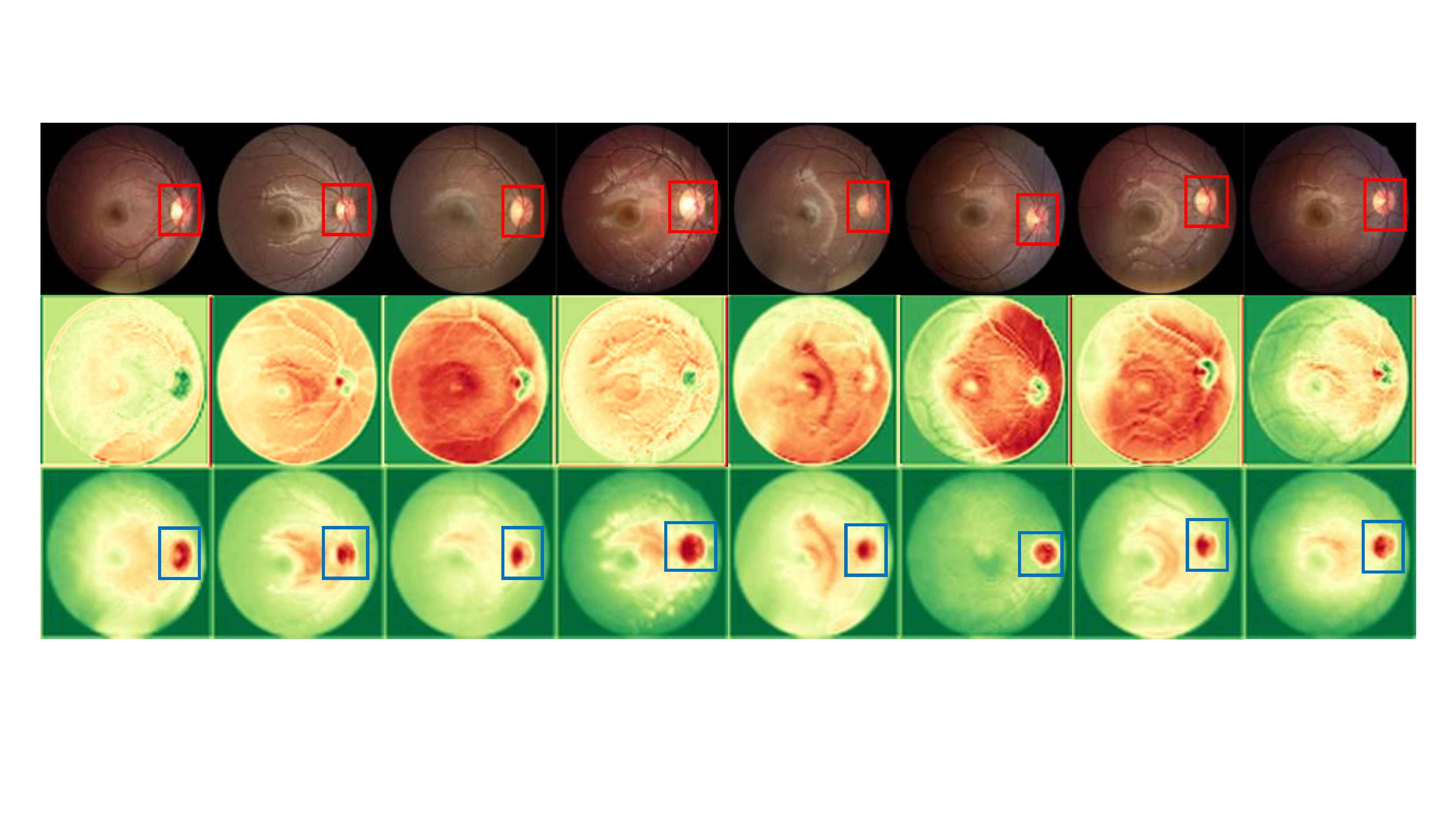}
\vspace{-1.8cm}
\caption{Visualization of learned feature maps by $\bm{s}$ and $\bm{z}$. The top row: original image; the middle row: feature maps of $\bm{z}$ by Grad-CAM; the bottom row: feature maps of $\bm{s}$ by Grad-CAM. The red to green corresponds to high to low response of the disease. As shown, the high response areas of $\bm{s}$ and $\bm{z}$ are respectively concentrated on the optic disc and other regions.} 
\label{fig:visualization}
\end{figure*}

\subsection{Ablation Study}
\label{sec:ablation}

In this section, we give a more comprehensive analysis regarding the effectiveness of the sequential modeling, clinical measurements provided as side information to help identification of latent space, and especially disentanglement of disease-causative hidden variables in handling OOD generalization. The compared variants are:

\textbf{Vanilla CNN}. Directly implement vanilla CNNs on images that share the same network structure with our encoder, and then we append two fully connected layers for prediction. The model does not make use of the time dependency between sequential data.

\textbf{CNN+LSTM}. We incorporate CNN into Long Short-Term Memory networks (LSTMs)~\cite{hochreiter1997long} 
to extract the features on sequential image data; then similarly, we use two fully connected layers for prediction. 

\textbf{Seq VAE}. We implement sequential variational autoencoder framework that shares the same time series architecture as ours. What's the difference from our method is that the clinical measurements $\bm{A}$ and personal attributes $\bm{B}$ are not provided and there is no disentanglement. 

\textbf{Seq VAE + Att}. The baseline is the same to ours only without separation of $\bm{s},\bm{v},\bm{z}$, to validate the advantage of disentanglement of disease-causative hidden variables $\bm{s},\bm{v}$ from others. 

\noindent \textbf{Results.} As shown in Tab.~\ref{tab:ablation}, the implementation of our time series model can achieve better results, showing the improvement of \textbf{Seq VAE} over \textbf{Vanilla CNN} and \textbf{CNN+LSTM}. The performance is further improved by leveraging the information of attributes (as shown by the improvement of \textbf{Seq VAE + Att} over \textbf{Seq VAE}). This improvement comes from two-fold contributions of attributes: \textit{i)} the observation of clinical measurements $\bm{A}$ that help identify the latent variable $v$; and \textit{ii)} the personal attributes $\bm{B}$ provided as auxiliary variables for learning the hidden variables. Finally, the improvement of \textbf{Ours} over \textbf{Seq VAE + Att} can validate the effectiveness of separating our disease-causative hidden variables in order to avoid spurious correlation.

\noindent \textbf{Robustness due to Disentanglement.} To further valid the benefit regarding the robustness of disease-causative hidden variables over others, we implement a two-step tuning method for $\bm{s},\bm{v}$ and $\bm{z}$. Specifically, after training the whole model at the first stage, we obtain the disease-unrelated hidden variable $\bm{z}$ and the disease-related hidden variables $\bm{s}$ and $\bm{v}$. We additionally train a classifier to predict the disease respectively by $\bm{s},\bm{v}$ and $\bm{z}$, \emph{i.e.,} $\bm{s},\bm{v} \to y$ and $\bm{z} \to y$. As shown in Tab.~\ref{tab:ablation_second_stage}, the $\bm{z}$ suffers from a significant performance drop from validation to test however the $\bm{s},\bm{v} \to y$ remains robust across validation and test. 
This can validate the existence of redundant variables that are spuriously correlated to the disease and can be learned during the data-fitting process. Separating these variables out during training can help avoid spurious correlation and therefore achieve more robustness on OOD samples.

\subsection{Visualization}
We visualize the high-response region for learned $\bm{s}$ and $\bm{z}$. Specifically, we implement Grad-CAM (Gradient-weighted Class Activation Mapping \cite{DBLP:journals/corr/SelvarajuDVCPB16}) to compute the feature map matrix by performing backpropagation on $\bm{s}$ and $\bm{z}$. The visualized feature maps are shown in Fig.~\ref{fig:visualization}, with the top row, the middle row, and the bottom row representing the original images, the visualized feature maps for $\bm{z}$ and $\bm{s}$. From red to green correspond to high to low response value. As shown, the high response area for $\bm{s}$ is concentrated to the optic disc as marked by blue rectangles, which was to be highly correlated to the disease status \cite{10.1167/iovs.12-9682,Savatovsky2015Longitudinal}. In contrast, the high-response area for $\bm{z}$ is scattered distributed into other areas such as macular. Due to data bias, the $\bm{z}$ can be spuriously correlated with the disease, however cannot generalize to other distributions.

\section{Conclusion and Future Work}
We propose a causal Hidden Markov Model for forecasting the disease at the future stage, given data up to the current stage. To enable OOD generalization on medical data that can suffer from a distributional change among populations, we propose to explicitly separate the disease-causative factors from others. Under the identifiable result that ensures the disentanglement of such disease-related hidden variables, we reformulate a new sequential VAE that conforms to our causal model for practical inference. The experimental results and the follow-up analysis validate the effectiveness and robustness of our model. The application of our method on more broad scenarios (such as Alzheimer's Disease) is left in our future work.

\section{Acknowledgements}
This work was supported by MOST-2018AAA0102004, NSFC-61625201 and NSFC-62061136001.


\newpage
{\small
\bibliographystyle{ieee_fullname}
\bibliography{egbib}
}

\end{document}